\documentclass{article} %
\usepackage{iclr2026_conference,times}

\usepackage{amsmath,amsfonts,bm}

\def\eqref#1{equation~\ref{#1}}

\def\1{\bm{1}}

\DeclareMathAlphabet{\mathsfit}{\encodingdefault}{\sfdefault}{m}{sl}
\SetMathAlphabet{\mathsfit}{bold}{\encodingdefault}{\sfdefault}{bx}{n}

\usepackage{hyperref}
\usepackage{url}
\usepackage{tabularx}
\usepackage[utf8]{inputenc} %
\usepackage[T1]{fontenc}    %
\usepackage{hyperref}       %
\usepackage{url}            %
\usepackage{booktabs}       %
\usepackage{amsfonts}       %
\usepackage{nicefrac}       %
\usepackage{microtype}      %
\usepackage{xcolor}         %
\usepackage[table]{xcolor}
\usepackage{algorithm}
\usepackage{algpseudocode}
\usepackage{graphicx}

\usepackage{multirow}
\usepackage{caption}
\usepackage{subcaption}
\usepackage{makecell}
\usepackage{csquotes}
\usepackage{epigraph}
\usepackage{pifont} %
\newcommand{\cmark}{\ding{51}} 
\newcommand{\xmark}{\ding{55}} 

\usepackage{placeins}
\usepackage[most]{tcolorbox}
\usepackage{arydshln}
\tcbset{colback=blue!5!white, colframe=blue!75!black, fonttitle=\bfseries}

\title{QueST: Incentivizing LLMs to Generate Difficult Problems}

\author{
\textbf{Hanxu Hu}$^{1}$\thanks{Work done during internship at Microsoft Research.} \quad \textbf{Xingxing Zhang}$^{2}$ \quad \textbf{Jannis Vamvas}$^{1}$ \quad \textbf{Rico Sennrich}$^{1}$  \quad \textbf{Furu Wei}$^{2}$ \\
\\
$^{1}$University of Zurich \quad $^{2}$Microsoft Research
}

\iclrfinalcopy %
\begin{document}

\maketitle

\begin{abstract}
\label{sec:abs}
Large Language Models have achieved strong performance on reasoning tasks, solving competition-level coding and math problems. However, their scalability is limited by human-labeled datasets and the lack of large-scale, challenging coding problem training data. Existing competitive coding datasets contain only thousands to tens of thousands of problems. Previous synthetic data generation methods rely on either augmenting existing instruction datasets or selecting challenging problems from human-labeled data. In this paper, we propose QueST, a novel framework which combines difficulty-aware graph sampling  and difficulty-aware rejection fine-tuning that directly optimizes specialized generators to create challenging coding problems. Our trained generators demonstrate superior capability compared to even \mbox{{\tt GPT-4o}} at creating challenging problems that benefit downstream performance. 
We leverage QueST to generate large-scale synthetic coding problems, which we then use to distill from strong teacher models with long chain-of-thought or to conduct reinforcement learning for smaller models, proving effective in both scenarios.
Our distillation experiments demonstrate significant performance gains. Specifically, after fine-tuning {\tt Qwen3-8B-base} on 100K difficult problems generated by QueST, we surpass the performance of the original {\tt Qwen3-8B} on LiveCodeBench. With an additional 112K examples (i.e., 28K human-written problems paired with multiple synthetic solutions), our 8B model matches the performance of the much larger {\tt DeepSeek-R1-671B}. These findings indicate that  generating complex problems via QueST offers an effective and scalable approach to advancing the frontiers of competitive coding and reasoning for large language models.

\end{abstract}

\vspace{-3ex}
\section{Introduction}
\label{sec:intro}

\begin{figure*}[htb]
\centering
\small
\vspace{4ex}
\centerline{\includegraphics[width=0.85\textwidth]{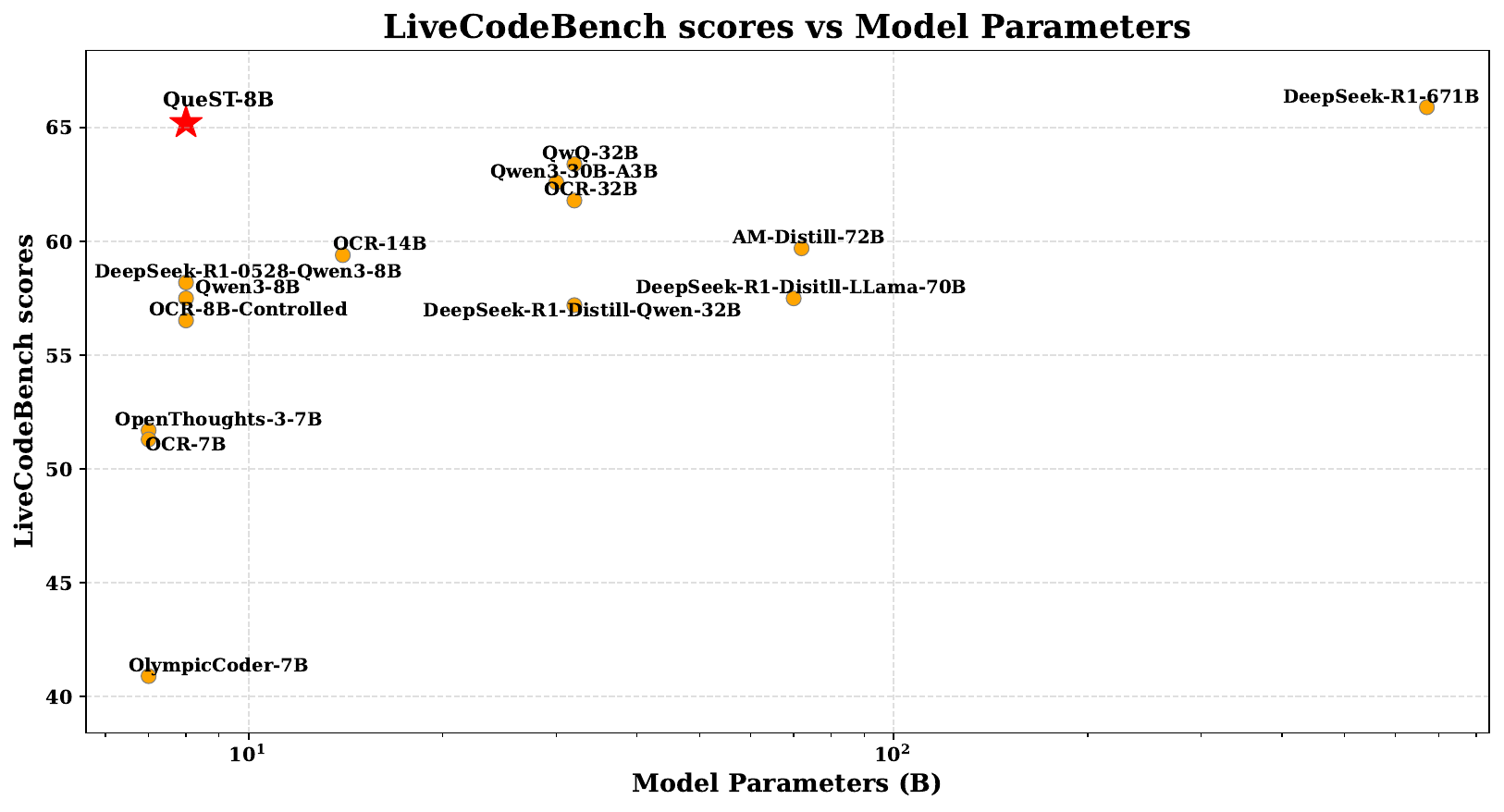}}
\caption{Comparisons of Livecodebench scores and model parameters between LLMs trained using various methods. Our model (QueST-8B)  achieves a new Pareto optimum.}
\vspace{1ex}
\label{fig:pareto}
\end{figure*}

Test-time scaling through long chain-of-thought and large-scale reinforcement learning has dramatically boosted the reasoning ability of large language models, enabling LLMs to solve competition-level coding and math problems that were previously beyond their reach. Models like OpenAI o1 \citep{O1} and DeepSeek-R1 \citep{R1} have demonstrated remarkable performance on challenging benchmarks such as Codeforces, AIME, and IOI, achieving expert-level problem-solving capabilities through extensive reasoning traces that can span thousands of tokens.
However, the problems used for training these models even require expert-level human annotations, which severely limits the scalability of LLM training. Current competitive coding datasets \citep{codecontest, taco} contain only thousands to tens of thousands of problems. As LLMs become more capable, the requirements for training data grow increasingly demanding—often requiring PhD-level experts in mathematics, computer science, and algorithm design to propose novel problems that can genuinely challenge these models. This process is not only extremely costly for requiring experts to propose difficult problems but also fundamentally cannot scale in terms of both dataset size and problem difficulty, creating a critical bottleneck in the development of next-generation reasoning models.

To mitigate this problem, methods for synthetic data generation and augmentation have been proposed. 
Previous works have focused either on paraphrasing-based augmentation \citep{luo2025wizardmath,yu2024metamath} or extracting concepts and recombining them based on co-occurrence probabilities \citep{mathscale, promptcot}. Some recent works have proposed leveraging model weaknesses and extracting concepts to create new problems \citep{liang2025swsselfawareweaknessdrivenproblem}.
More recently, reasoning-based LLMs have presented the next paradigm in advancing large language model reasoning capabilities \citep{O1, R1}. Works like \citet{R1, guha2025openthoughtsdatarecipesreasoning} created long chain-of-thought responses from reasoning models and curated synthetic SFT datasets, effectively helping small open-weight LLMs achieve superior performance in code and math tasks. \cite{OCR} curated the largest open-source dataset by obtaining long CoT responses from DeepSeek-R1 multiple times for each problems, though the problems themselves are still sourced from human-labeled competition coding problems. These methods have narrowed the performance gap between open-weight models and closed-weight reasoning models.

However, existing methods described above either focus on leveraging existing human-annotated problems and curating synthetic responses from existing reasoning models, or rely on a \emph{fixed} LLM %
to generate new problems by prompting. In this paper, we are the first to propose a method that directly trains an LLM generator to create challenging competitive code reasoning problems. We call our method \textbf{QueST}, embarking on a quest to generate increasingly challenging code problems through the combination of difficulty-aware graph sampling and difficulty-aware rejection fine-tuning. This approach is more scalable and flexible compared to previous methods that used a fixed generator or fixed human-labeled problems. Our proposed method makes the generator specialized and stronger than even proprietary models in creating problems that are challenging and useful for training of downstream tasks. We leverage this to generate the largest-scale code problem training set compared to previous synthetic data approaches, and the statistics of our synthetic data with previous data are shown in Table \ref{tab:compare_data}. We obtained responses from long chain-of-thought reasoning models, then leveraged the generated datasets to SFT small models. 
As illustrated in Figure \ref{fig:pareto}, our 8B model (QueST-8B), trained on a combination of 100K QueST-generated examples and an additional 112K examples derived from human-written problems, achieves state-of-the-art performance among models of similar size on code reasoning benchmarks. Notably, its results closely approach those of the much larger DeepSeek-R1 671B.

Our contributions can be summarized as follows:
\begin{itemize}
    \item We introduce a novel difficulty-aware coding problems generation framework that combines both difficulty-aware graph sampling and difficulty-aware rejection fine-tuning, which trains specialized generators to create challenging coding problems.
    \item We create the largest synthetic code reasoning problem set to date, comprising over 100K challenging coding problems paired with detailed chain-of-thought solutions from reasoning models.
    \item We train 8B base model using our synthetic data combined with original data and achieved state-of-the-art performance with only 212K samples. Our model reaches the new Pareto optimum as shown in Figure \ref{fig:pareto}.
    \item We conduct comprehensive ablation studies and analyses of our proposed method and the distribution of the generated coding problems.
\end{itemize}

\begin{table}[]
\centering
\caption{Comparison between representative code reasoning datasets.}
\small
\begin{tabular}{@{}llcc@{}}
\toprule
Code Datasets                & \textbf{\#Problems} & Long CoT Responses & Synthetic Problems \\ \midrule
CodeContest \citep{codecontest}             & 13K                  &       \xmark                    &           \xmark            \\
TACO \citep{taco}                   & 26K                  &      \xmark                           &   \xmark                          \\
Bespoke-Stratos \citep{bespoke_stratos}         & 17K                  &       \cmark                          &    \xmark                      \\
Open-R1 Codeforces-cots \citep{openr1} & 10K                  &       \cmark                          &   \xmark                          \\
OpenCodeReasoning \citep{OCR}      & 28K                  &        \cmark                         &      \xmark                       \\
\textbf{Ours (QueST)}            & \textbf{100K}        &    \cmark                              &    \cmark                         \\ \bottomrule
\end{tabular}
\label{tab:compare_data}
\end{table}

\section{QueST}
In this section, we present \textsc{QueST}, our proposed method for generating difficult problems. We focus our investigation on the generation of coding problems, as other forms of reasoning tasks (e.g., mathematical reasoning) can be regarded as special cases of coding tasks \citep{pfpo}.
We begin by introducing our scaffolding framework for problem generation, which builds upon MathScale \citep{mathscale}. Next, we detail our strategies for incentivizing LLMs to produce more difficult problems. Finally, we demonstrate how our scaffolding can be adapted to further enhance the generation of challenging problems.

\subsection{Preliminary: Problem Generation Through Concept Graph}
\label{sec:q_gen_concept}

Our scaffolding for problem generation is based on \citep{mathscale}, which generates new problems based on existing seed problems by prompting an LLM in three steps (i.e., concept extraction, graph construction and problem generation).

\paragraph{Concept Extraction} For each problem $q$ in the seed problem set $\mathbf{Q}_{\text{seed}}$, we prompt an LLM to extract concepts $c$ (topics and knowledge points) from it. We follow the setting of \cite{mathscale}: \textit{Topics} refers to general directions, \textit{knowledge points} refers to more fine-grained concepts; an example can be found in Appendix Table \ref{fig:promptgen}. Note that problem generation can be guided later using the concepts that we extracted in this step. The process can be defined formally as follows:
\begin{equation}
    \mathbf{C} = \mathcal{G}^{\text{Q}}((p_{\text{extract}}, \mathbf{Q}_{\text{seed}}))
\end{equation}
where $p_{\text{extract}}$ is the prompt used to extract concepts, $\mathcal{G}^{\text{Q}}$ is our problem generator and $\mathbf{C}$ is the set of concepts extracted. Detail prompts of $p_{\text{extract}}$ are in Appendix Table \ref{fig:promptconcept}.

\paragraph{Graph Construction} 
Once we obtain the concepts $\mathbf{C}$, we proceed to identify plausible combinations of these concepts. Two concepts are considered to form a reasonable combination if they have frequently co-occurred within the same problem in the seed dataset. To capture these relationships, we construct a concept graph in which nodes represent individual concepts, and edge weights encode the strength of co-occurrence between concept pairs. The edge weight $w(u, v)$ is defined as follows:
\begin{equation}
\label{eq:edge_weight}
w(u, v) = \log\left(\mathrm{freq}(u, v) + \epsilon\right)
\end{equation}
where $u$ and $v$ denote concept nodes, and $\mathrm{freq}(u, v)$ represents the observed co-occurrence frequency of these concepts. A small constant $\epsilon$ is added to ensure numerical stability by preventing zero counts.

Given the constructed graph, we proceed to sample concept combinations, which are then utilized for the generation of new problems. We start from a uniformly random sampling from all the topics and subsequently perform up to six steps of a random walk on the graph \citep{mathscale}. At each step, the transition probability from node $\mathbf{u}$ to node $\mathbf{v}$ is defined as:
\begin{equation}
p_{\mathbf{u}, \mathbf{v}} = \frac{\exp\left(w(\mathbf{u}, \mathbf{v})\right)}{\sum_{\mathbf{v}' \in \mathcal{N}(\mathbf{u})} \exp\left(w(\mathbf{u}, \mathbf{v}')\right)}
\end{equation}
where $\mathcal{N}(\mathbf{u})$ denotes the set of nodes adjacent to u.
After each random walk episode, we obtain a sampled concept combination $s$, which is subsequently used for problem generation.

\paragraph{Problem Generation} 

Given the sampled concept set $s$, we leverage an LLM to generate new problems. We incorporate few-shot examples to guide the LLM in formulating problems. These examples are selected from the pool of seed problems based on the Jaccard distance between their respective sets of concepts. Formally, this process can be described as:
\begin{equation}
\label{eq:gen_q}
\mathbf{Q}_{\text{new}} = \mathcal{G}^{\text{Q}}(p_{\text{generate}}, \mathcal{S}(\mathbf{C}), \mathbf{Q}_{\text{seed}})
\end{equation}
where $\mathcal{S}(\mathbf{C})$ denotes the set of sampled concepts, and $p_{\text{generate}}$ represents the prompt template utilized for problem generation (see Appendix Table \ref{fig:promptgen} for additional details).

At this stage, our problem generator is designed to produce new problems, rather than explicitly targeting increased difficulty. The generation of more challenging problems will be addressed in the subsequent sections.

\subsection{Difficulty-aware Rejection Finetuning}

\begin{figure*}[t]
\centering
\vspace{4ex}
\centerline{\includegraphics[width=\textwidth]{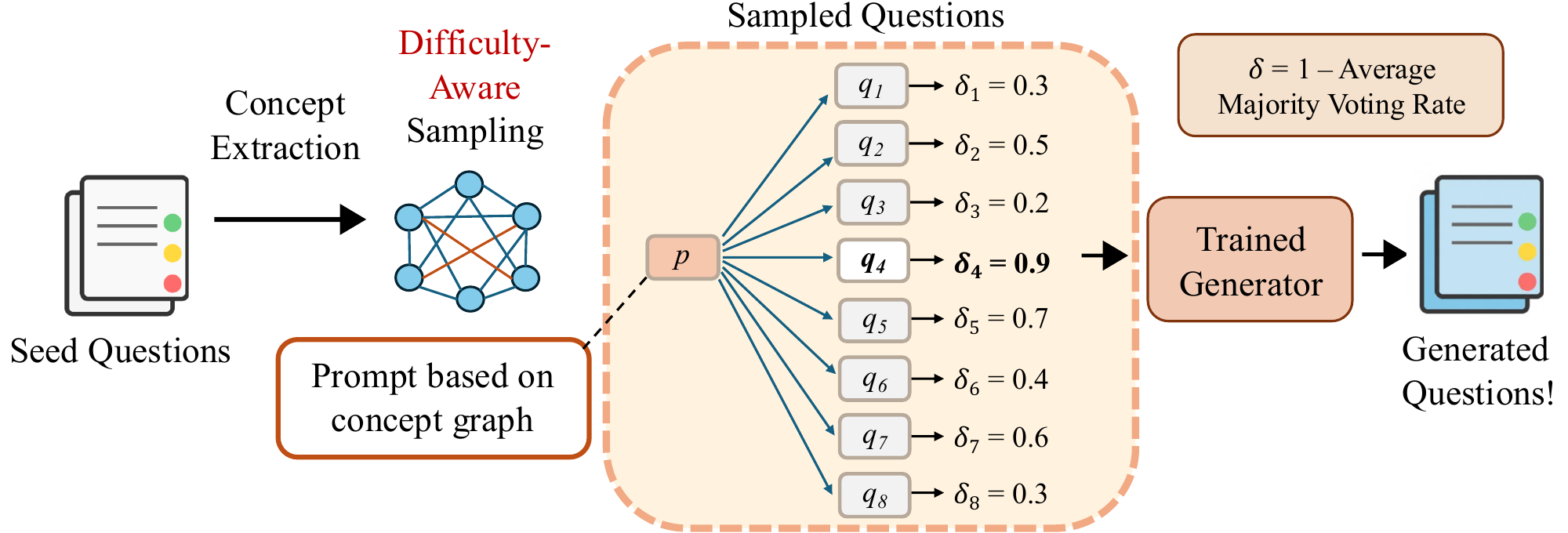}}
\caption{The pipeline of QueST. We first extract concepts based on seed problems, then use difficulty-aware sampling method described in Equation \ref{eq:edge-weight-diff} to create prompts for problem generation. We generate 8 problems for each prompt, calculate the difficulty $\delta$ of the generated problem based on Equation \ref{eq:diff}, and use the most difficult problem as rejection fine-tuning data to train our generator. }
\vspace{1ex}
\label{fig:mainfig}
\end{figure*}

\label{sec:rft_gen}

We focus primarily on the generation of challenging coding problems, though our approach is readily extensible to other forms of reasoning tasks. We first present our method for measuring problem difficulty, and then illustrate how this measure is employed to guide LLMs in producing more difficult problems.

\paragraph{Difficulty Estimation} 
\label{sec:diff_est}

A natural way to assess the difficulty of a generated problem is to examine the consistency of the models multiple outputs. \cite{wang2023selfconsistency} finds that self-consistency is highly correlated with accuracy, which reflects the uncertainty of the models, and also the difficulty of the problem.
When most solutions converge to a single outcome, the problem is likely straightforward. Conversely, if the solutions diverge and produce inconsistent outputs, this indicates model uncertainty, suggesting the problem is more difficult. 

Building on this intuition, we define the difficulty of a problem using the \emph{average majority voting rate} of its solutions. We illustrate our metric using coding problems as a case study, noting that other verifiable reasoning problems (e.g. math) can be regarded as a special case of this setting. Let $q \sim \mathcal{G}^{\text{Q}}(p_{\text{generate}}, s, \mathbf{Q}_{\text{seed}})$
denote a generated coding problem (see Equation~(\ref{eq:gen_q})), where $s \sim \mathcal{S}(\mathbf{C})$ is a sampled concept combination (see Section \ref{sec:q_gen_concept}).
The estimation proceeds in three steps. First, we prompt {\tt gpt-4o} to 
generate $T$ test inputs, forming the set $\mathcal{I} = \{i_1, i_2, \ldots, i_T\}$ (details of the prompt in Appendix Figure \ref{fig:prompttestcase}). Second, we obtain $M$ candidate solutions $\mathcal{Y} = \{ y_1, y_2, \ldots, y_{M} \}$ from {\tt gpt-4o}. Third, we execute each $y_m \in \mathcal{Y}$ on all inputs $i_t \in \mathcal{I}$, producing output sets $\mathcal{O}_t = \{ g(y_1, i_t), g(y_2, i_t), \ldots, g(y_M, i_t) \}$, where $g(y_m, i_t)$ denotes extracting the code from $y_m$, running it on input $i_t$, and recording the output. For each test input, the most likely output $o_t$ is identified as the most frequent element in $\mathcal{O}_t$:
\begin{equation}
    o_t = \arg\max_{o \in \mathcal{O}_t} f(o, \mathcal{O}_t)
\end{equation}
where $f(o, \mathcal{O}_t) = \bigl| \{ x \in \mathcal{O}_t \mid x = o \} \bigr|$ counts the occurrences of $o$ in $\mathcal{O}_t$. Finally, we quantify the problem difficulty as
\begin{equation}
    \delta(q) = 1 - \frac{1}{T} \sum_{t=1}^T \frac{f(o_t, \mathcal{O}_t)}{M}
    \label{eq:diff}
\end{equation}
Intuitively, $\delta(q)$ measures the degree of disagreement among candidate solutions: the lower the majority voting rate, the higher the difficulty. Thus, larger values of $\delta(q)$ correspond to more challenging problems.

To further enhance the probability of generating valid synthetic problems, we filter out problems where over half of the test case outputs from generated responses return None, indicating unsuccessful code execution. 
\paragraph{Rejection Fine-tuning}

Having introduced the difficulty measure $\delta(q)$, we now describe how it is employed to construct a dataset of prompt–problem pairs for training LLMs to generate difficult problems. The key idea is to sample multiple candidate problems from the same prompt and retain only the most difficult one.

As discussed in Section~\ref{sec:q_gen_concept}, for each concept combination $s$, a problem can be generated via
\begin{equation}
    q \sim \mathcal{G}^{\text{Q}}(p_{\text{generate}}, s, \mathbf{Q}_{\text{seed}})
\end{equation}
where $\mathcal{G}^{\text{Q}}$ denotes the LLM-based generator. More generally, let $M_{\theta}$ be the LLM parameterized by $\theta$, and let $p$ denote the \emph{actual} prompt ($p_{\text{generate}}$ instantiated with concept set $s$ and seed problems $\mathbf{Q}_{\text{seed}}$) used to query $M_{\theta}$. By sampling $K$ times, we obtain a set of candidate problems:
\begin{equation}
    q_k \sim M_{\theta}(p) \quad \text{for } k = 1, \dots, K
\end{equation}
We denote this set by $\mathcal{Q} = \{q_1, q_2, \dots, q_K\}$. We then select the most difficult problem according to our measure $\delta(\cdot)$:
\begin{equation}
q^* = \arg\max_{q_k \in \mathcal{Q}} \delta(q_k)
\end{equation}
Only $q^*$ is retained, while the remaining candidates are discarded. The resulting pair $(p, q^*)$ is added to the training set $\mathcal{D}_{\text{hard}}$, which is used to fine-tune the problem generator $M_{\theta}$.

\subsection{Difficulty-aware Graph Construction}
\label{sec: diff_aware}
This section extends our problem generation scaffolding (Section~\ref{sec:q_gen_concept}) to be difficulty-aware. In the baseline setup, the initial edge weights of the concept graph are determined primarily by the co-occurrence statistics of concepts within the same problems. Here, we further incorporate difficulty by modeling the hardness of concepts with respect to the difficulty levels of the problems in which they appear.
Since each problem in the seed dataset (e.g., TACO; ~\cite{taco}) is annotated with human-curated difficulty labels, we leverage this information when constructing the concept graph for problem generation prompts. Specifically, beyond using co-occurrence counts as edge weights for random walk sampling, we also incorporate the average difficulty of all problems that involve both concepts connected by an edge. The new edge weights are defined as
\begin{equation}\label{eq:edge-weight-diff}
\begin{aligned}
w(u, v) &= \log\left(\alpha \cdot \mathrm{freq}(u, v) + \textcolor{red}{ (1 - \alpha) \cdot \mathrm{diff}(u, v)} + \epsilon\right)
\\
&\text{where }\textcolor{red}{\mathrm{diff}(u, v)
= \frac{1}{\lvert Q_{u,v} \rvert} \sum_{q \in Q_{u,v}} d(q)},\quad
Q_{u,v} = \{\, q \mid u \in q,\ v \in q \,\}.
\end{aligned}
\end{equation}
Here, $\alpha$ is a hyperparameter that balances the contribution of co-occurrence frequency and difficulty and we set $\alpha=0.2$ in our experiments. 
The constant $\epsilon$ is included to avoid taking the logarithm of zero.
The set $Q_{u,v}$ consists of all problems containing both concepts $u$ and $v$; its cardinality is denoted by 
$|Q_{u,v}|$.
Finally, 
$d(q)$ represents the human-annotated difficulty level of problem 
$q$, given as an integer from 1 to 5.

\section{Experiments}
In this section, we present the detail of our experiments. We first use our proposed difficulty measure method for data selection. Then we show the long CoT SFT results using datasets distilled from Qwen3-8B compared with previous strong baselines, and we show our generated datasets can also be effective when used in RL training. We further present an ablation study to investigate the effect of each role in our proposed method. Finally we have contamination analysis and statistics about our generated data.

\subsection{Implementation Details}
\label{sec:implement}
\textbf{Seed data:} We use TACO \citep{taco} as seed data, which has human-annotated labels for difficult. TACO has 25.4K training samples and 1K test samples. Each problems is annotated with difficulty,  test cases, and a list of topics. Samples in this dataset are collected from open-access sites where programmers share problems with each other, including Aizu, AtCoder, CodeChef, Codeforces, and LeetCode.

\textbf{Benchmarks}
We use LiveCodeBench-V5 \citep{LCB} and USACO as our evaluation benchmarks. We use LiveCodeBench-V5 for direct comparison with a strong baseline \citep{OCR}; USACO \citep{USACO} is used because it is a representative code competition which contains difficult problems and has already been curated as benchmark for evaluation.  

\textbf{Models:} We use Qwen3-8B as our teacher model in distillation experiments, as it is efficient and has competitive reasoning performance. For the final large scale distillation experiments (Table \ref{tab:performance}), we employ Qwen3-235B-A22B \citep{yang2025qwen3} as our teacher. We use Qwen2.5-Coder-7B-Instruct \citep{hui2024qwen2} and Qwen3-8B-Base as our student model, respectively. For the RL experiments, we use Qwen2.5-7B-Instruct \citep{yang2025qwen3} model  as starting checkpoint for small-scale verification. We use Qwen2.5-14B-Instruct and GPT-4o as generators, as they can follow instructions relatively well compared to smaller models.

\textbf{Hyperparameters:}  We use vLLM \footnote{\url{https://github.com/vllm-project/vllm}} as our inference framework for both distillation and evaluation experiments. We set temperature to 0.6 for all experiments. We set the batch size to 128 and the learning rate to 5e-5 for our SFT experiments, including the fine-tuning of the generator models. We use VeRL \footnote{\url{https://github.com/volcengine/verl}} for our RL experiments,  and use 128 as the rollout batch size, 64 as the mini-batch size, and 16 as the rollout sample size. For all evaluation, we calculate averaged pass@1 across 16 runs.

\subsection{Using Estimated Difficulty For Data Selection}
\label{sec:selection}
Before training the generator to produce difficult coding problems, we first need a trustworthy signal that can serve as a proxy for difficulty when gold labels are unavailable for generated problems. As mentioned above, we propose using $\delta$ we defined in Section \ref{sec:diff_est} based on model responses.
To verify the usefulness of this signal, we conduct a preliminary experiment that selects subsets of generated problems based on this signal for controlled comparison.
We use our baseline graph random walking process to generate 50K problems using TACO as seed data. For each problem, we generate 8 responses and compute $\delta$. We then select 3K samples with the highest $\delta$, 3K with the lowest $\delta$, 3K with $\delta$ closest to 0.5, and an additional 3K randomly sampled for comparison. We also use response token length as another difficulty proxy and select 3K samples with the longest responses. Table \ref{tab:select} shows the results of using different selection methods and the performance of models trained on the selected problems, with 8 responses generated for each problem to ensure the scale and significance of our experiments. We observe that problems with the highest $\delta$ achieve the best performance, even surpassing those with the longest token responses, and using significant less tokens. We can also observe that for the problems with highest $\delta$, the token length is higher than problems with median and lowest $\delta$, which indicates there are some positive correlations between token length and $\delta$, but $\delta$ is still a more effective and efficient signal compared to response length.
\begin{table}[]
\centering
\caption{Effect of different strata of synthetically generated coding problems on downstream performance. $\delta$ refers our estimated difficulty defined in Section \ref{sec:diff_est} . Response length is determined based on responses generated by Qwen3-8B.}
\begin{tabular}{@{}lrr@{}}
\toprule
Selection of problems & LiveCodeBench-V5 score & Avg. response length in tokens \\ \midrule
Random 3K               &     36.29        &    11.9K                   \\
\rowcolor{blue!10} Highest $\delta$ 3K            &    \textbf{39.28}         &   14.2K                    \\
Median $\delta$ 3K            &    36.35          &  14.1K                     \\
Lowest $\delta$ 3K           &     32.37        &   6.8K                    \\
Longest response 3K        &   38.35          &    22.6K                \\ \bottomrule
\end{tabular}
\label{tab:select}
\end{table}

\subsection{Trained Generator for Distillation}
We then use our trained generator to generate problems and leverage these problems to obtain responses from long chain-of-thought models (Qwen3-235B-A22B in our experiments) for training student models. In Table \ref{tab:main}, we conduct a comprehensive comparison between previous Long CoT SFT datasets and our generated datasets on representative code reasoning benchmarks (i.e., LiveCodeBench-V5 and USACO). 
As detailed in Section~\ref{sec:implement}, we employ {\tt Qwen2.5-14B-Instruct} to train a specialized generator using our rejection fine-tuning and difficulty-aware graph sampling techniques. This approach enables us to generate a set of 100K challenging coding problems, each paired with a response from Qwen3-235B-A22B.
We adopt Qwen3-8B-Base as the student model and benchmark its performance against other models of comparable size trained on prior synthetic datasets (see the second block in Table \ref{tab:performance}). Our model, QueST-100K-8B, consistently surpasses all similarly sized baselines. Among the comparison group, OCR-Qwen-7B-Instruct \citep{OCR} stands out as the strongest competitor, leveraging DeepSeek-R1 as the teacher model and generating up to 32 responses for each of the 28K human-written coding problems. To ensure a fair comparison, we re-implement the OCR method using Qwen3-8B-Base as the student model and generate 4 responses per problem (yielding a total of 112K examples) using Qwen3-235B-A22B as the teacher (see OCR-8B in the third block). Even under these conditions, QueST-100K-8B outperforms OCR-8B across both benchmarks, suggesting that a large volume of diverse, challenging synthetic problems provides greater benefit than simply repeating existing human-written ones. Furthermore, by combining the 112K organic examples with our 100K generated examples, we achieve even stronger results (QueST-8B), attaining performance on par with the much larger DeepSeek-R1-671B.

\begin{table}[ht]
  \centering
  \small
  \caption{Performance on LiveCodeBench-V5 (from 2409 to 2502) and USACO. Note: We use Qwen3-235B-A22B as the teacher model and Qwen3-8B-Base as the student model. 
  In our methods, we use our trained Qwen2.5-14B-Instruct as generator. 100K means the number of training samples.}
  \label{tab:performance}
  \resizebox{0.95\textwidth}{!}{
  \begin{tabular}{@{}lrrrrrrrr@{}}
    \toprule
    Model
      & \multicolumn{4}{c}{LiveCodeBench-V5}
      & \multicolumn{4}{c}{USACO} \\
    \cmidrule(lr){2-5} \cmidrule(lr){6-9}
      & Easy & Medium & Hard & Avg.
      & Easy & Medium & Hard & Avg. \\
    \midrule   
    DeepSeek‐R1-671B \citep{R1}      & 98.5 & 79.8 & 37.4 & 65.6 & 72.5  & 54.6 & 34.3 & 56.2 \\
    \midrule                                   %
    Qwen3-8B \citep{yang2025qwen3}         & 94.0 & 74.1 & 28.9 & 58.7 & 58.5  & 42.8 & 22.3 & 43.5  \\%
R1-0528-Qwen3-8B \citep{R1} & 94.4 & 73.5 & 27.7 & 58.1 &57.0  &33.6 &17.2 & 38.5 \\
    OpenThinker‐7B (114K) \citep{guha2025openthoughtsdatarecipesreasoning}      & 80.6 & 16.9 &  1.6 & 25.5 & 11.0 & 2.1 & 0.0  &  5.0 \\
    R1‐Distill‐Qwen‐7B (800K) \citep{R1} & 86.6 & 43.8 &  7.0 & 38.0 & 22.9 & 9.7 & 3.8  & 13.4 \\
    OlympicCoder‐7B (100K) \citep{openr1}     & 82.1 & 49.4 & 12.2 & 40.9 & 31.4 & 12.5 & 1.3 & 17.0 \\
    OCR‐Qwen‐7B‐Instruct (700K) \citep{OCR} & 95.4 & 64.0 & 18.0 & 51.3 & 41.5  & 26.0 & 7.5 & 27.2 \\
    \midrule 
    OCR-8B (112K, Our Impl.)   & 96.0 & 70.2 &  26.2  & 56.5 &  54.5 &  40.5 & 22.9 & 41.3 \\                                   %
    QueST-100K-8B (100K, Ours)  & 97.1  & 74.8  & 28.4   & 59.4 & 55.9  & 44.0 & 24.7  & 43.5  \\
    \rowcolor{blue!10} QueST-8B (212K, Ours)  & \textbf{97.6} & \textbf{81.0} &  \textbf{36.6}  & \textbf{65.2} & \textbf{65.5} & \textbf{48.6}  & \textbf{28.7} & \textbf{49.9} \\
    \bottomrule
  \end{tabular}
  }
\label{tab:main}
\end{table}

\subsection{Reinforcement Learning}
Our generated data can also be used for RLVR (Reinforcement Learning with Verifiable Reward). We use majority voting results produced by Qwen3-8B as pseudo output labels for each test case of each generated problem. Since our generated test cases are not guaranteed to be valid, we filter out test cases where over half of the outputs are none (indicating failed execution for generated solutions), then keep the remainder for RLVR. We use the GRPO \citep{deepseekmath} algorithm to train Qwen2.5-7B-Instruct on 12K problems sampled from TACO, 6K data from our baseline synthetic method (mathscale) \citep{mathscale}, and 6K data from QueST. We report our results in Table \ref{tab:rl}, which shows effectiveness of our proposed method.

\begin{figure*}[t]
\centering
\small
\vspace{4ex}
\centerline{\includegraphics[width=0.7\textwidth]{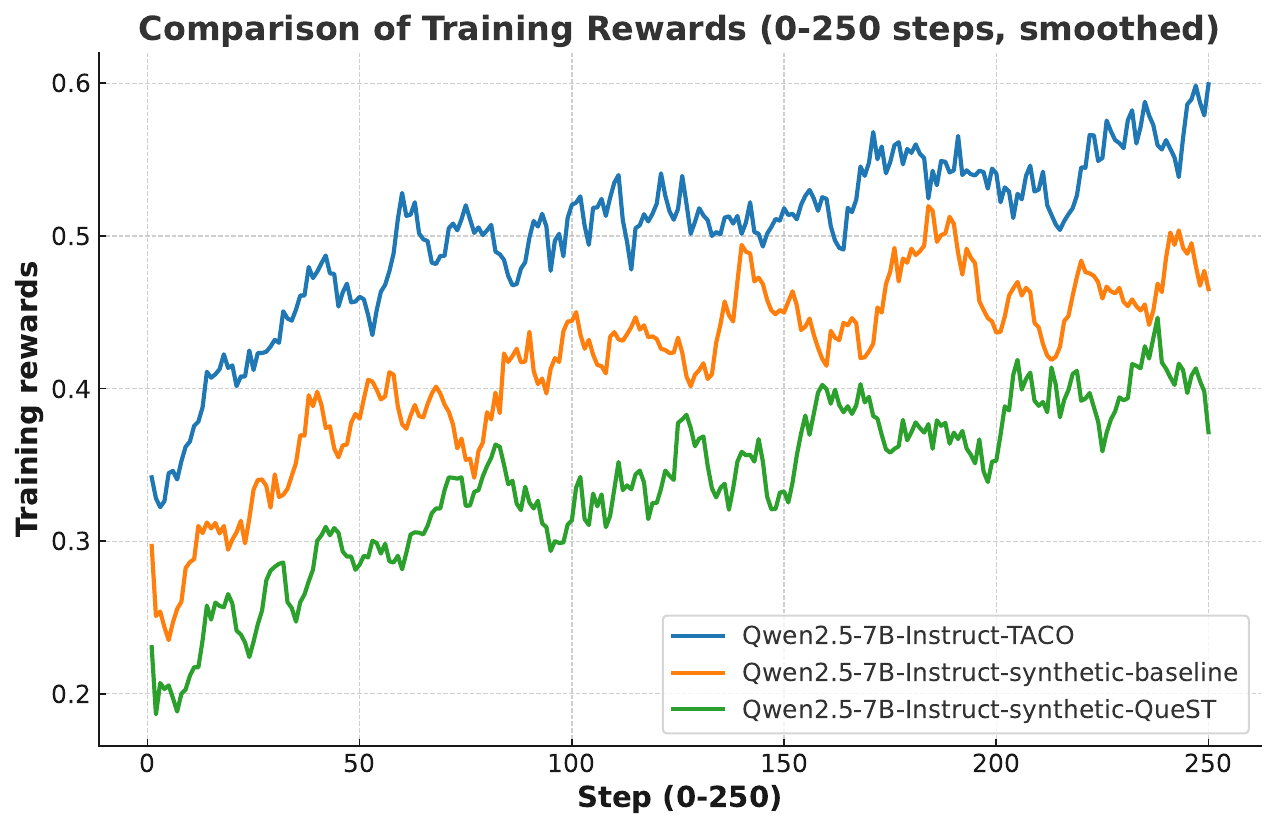}}
\caption{Training rewards comparison in the training process of RL under different datasets.}
\vspace{1ex}
\label{fig:rewards}
\end{figure*}

We report the training reward curve during the training process in Figure \ref{fig:rewards}. It shows that the model trained on TACO datasets gains the highest reward score during the whole training stage, our baseline synthetic method gains a lower score, and the model trained on a dataset generated by the QueST method gains the lowest score. The training reward can serve as a proxy of the inverse difficulty of these three different datasets.

\begin{table}[]
\centering
\small
\caption{RL results on LiveCodeBench-V5}
\begin{tabular}{@{}lrrrr@{}}
\toprule
\multirow{2}{*}{Model}           & \multicolumn{4}{c}{LiveCodeBench-V5} \\ \cmidrule(l){2-5}
& Easy   & Medium   & Hard  & Avg.  \\ \midrule
Qwen2.5-7B-Instruct              &    47.4    &    8.4      &    0.1   &   14.3  \\
Qwen2.5-7B-Instruct TACO RL &   56.7     &   10.8     &   1.1    &  17.3    \\
Qwen2.5-7B-Instruct Baseline RL &   56.0     &  9.6        &  3.2     &  17.6     \\
\rowcolor{blue!10} Qwen2.5-7B-Instruct QueST RL     &   56.4     &    9.6      &  4.8     &  18.6     \\ \bottomrule
\end{tabular}
\label{tab:rl}
\end{table}

\subsection{Ablation study}
We conducted an ablation study for fair comparison across different settings, as shown in Table \ref{tab:ablation}. In the first two rows of the table, we examine whether using difficulty-aware graph random walking improves performance when using GPT-4o as the generator. The results demonstrate that the difficulty-aware graph achieves clear improvement. In the third and fourth rows, we compare performance when using the difficulty-aware graph with different generators: Qwen2.5-14B-Instruct without further training and Qwen2.5-14B-Instruct under our rejection fine-tuning method (QueST). The results show that when using difficulty-aware random sampling prompts, our fine-tuned generator can bring better performance than the model without using our fine-tuning method. Therefore, Table \ref{tab:ablation} indicate both difficulty-aware sampling and rejection fine-tuning have positive effect and lead to generating difficult problems. 
We conducted controlled comparisons using stronger teacher model (Qwen3-235B-A22B) and base model (Qwen3-8B-Base) at larger scale in Table \ref{tab:abl_large}. 
Models trained solely on our synthetic problems achieve higher performance than those trained exclusively on OCR problems, albeit with significantly longer response lengths, indicating that our generated problems are more challenging. In contrast, combining OCR problems with our synthetic problems yields the best overall performance while also reducing response lengths.

\begin{table}[ht]
  \centering
  \small
  \caption{Ablation study on LiveCodeBench-V5. ``Baseline'' here represents the our baseline problem generation pipeline \citep{mathscale} which we discussed in Section \ref{sec:q_gen_concept}. Here we generate 20K questions for all settings to fair comparison, and the base model we used to train is Qwen2.5-Coder-7B-Instruct. ``RFT'' is abbreviation of our rejection fine-tuning method.}
  \label{tab:ablation}
  \begin{tabular}{@{}lrrrr@{}}
    \toprule
    Methods & \multicolumn{4}{c}{LiveCodeBench-V5} \\
    \cmidrule(lr){2-5}
          & Easy & Medium & Hard & Avg. \\
    \midrule
    \multicolumn{5}{c}{Problem Generator: GPT-4o} \\
    \midrule
    Baseline                & 82.8 & 36.6 &  8.2 & 34.9 \\
    Baseline w/ difficulty-aware graph       & 83.6  & 41.1 & 10.9  & 37.5 \\
        \midrule
    \multicolumn{5}{c}{Problem Generator Qwen2.5-14B-Instruct} \\
     \midrule
 Baseline w/ difficulty-aware graph        & 85.0 & 39.2  & 8.0  & 36.1 \\
   Baseline w/ difficulty-aware graph w/ RFT (QueST)       & 84.9 & 41.4 & 10.4 & 37.9 \\
    \bottomrule
  \end{tabular}
\end{table}

\vspace{-3mm}

\begin{table*}[h]
\centering
\small
\caption{Performance and average response length of models trained on our 100K synthetic problems, 112K OCR problems, and merged one (100K + 112K). }
 \label{tab:abl_large}
\begin{tabular}{@{}llll@{}}
\toprule
Models             & QueST (merged) & QueST (synthetic only) & OCR 112K \\ \midrule
Performance on LCB & \textbf{65.2}              & 59.4       & 56.5     \\
Avg. response length in tokens      & 30503             & 34347      & \textbf{22969}    \\ \bottomrule
\end{tabular}
\end{table*}

\subsection{Additional Analysis}
We visualize and compare the 25 most sampled knowledge points with and without difficulty-aware sampling in Appendix Figure \ref{fig:freq}. The figure shows that knowledge points sampled more frequently by naive sampling than by difficulty-aware sampling tend to be more common overall, while knowledge points sampled less frequently by naive sampling tend to be less common. In other words, difficulty-aware sampling upweights infrequent knowledge points and downweights frequent knowledge points compared to naive sampling. The infrequent knowledge points are visualized in the left figure and are generally more difficult, including topics such as the ``knapsack problem'', ``Optimal Play Strategies'', and ``prime factorization'', compared to the basic concepts shown in the right figure.

We also conduct a case study on generated problems from both original model and model trained by QueST framework in Appendix Table \ref{tab:case-study}. It shows that the problem generated by our trained model requires more complex operations and more knowledge compared the question generated by original model.

To prove that our results are not affected by data contamination, we conduct contamination detection experiments on our generated datasets to exclude the effects of data contamination on benchmark performance. Specifically, we compute token-based 50-gram Jaccard similarity scores and the scores across all datasets and benchmarks we used are 0 which indicates there is no contamination in our generated data.

\section{Related Work}
\subsection{Synthetic Data for Language Models}
Synthetic data has been widely used in training language models. Previous works have mainly focused on using small sets of seed data and leveraging LLMs to augment them and generate larger datasets. Some works \citep{honovich-etal-2023-unnatural, li2024common7blanguagemodels, toshniwal2025openmathinstruct, wang-etal-2023-self-instruct, mathscale} focus on sampling seed data as in-context learning exemplars to generate new ones. \citet{ge2025scalingsyntheticdatacreation} proposed using personas to augment previous in-context learning synthetic data generation methods. \citet{xu2024wizardlm, luo2025wizardmath, hu-etal-2025-fine-tuning} focus on augmenting existing samples to create more complex ones. Some methods have also explored how to generate synthetic data from scratch \citep{li2024syntheticdataalmostscratch, xu2025magpie}. More recently, \citet{qin2025scaling} investigated whether synthetic data follows similar scaling laws as real data. PromptCoT \citep{promptcot} also generates challenging problems based on mathematical concepts and rationale. \citet{tong2024dartmath} also proposed a difficulty-aware method but focuses on synthetic responses for challenging problems. \citet{liang2025swsselfawareweaknessdrivenproblem} extract concepts from failure cases and synthesize new problems during RL training. Additionally, there is research focused on leveraging pretraining or web data to generate reasoning data in general domains \citep{yuan2025naturalreasoningreasoningwild28m, yue2024mammoth2}. Our QueST framework focuses on a new perspective that aims to train a difficulty-aware generator to generate difficult problems.
\vspace{-3mm}

\subsection{Code Reasoning}

Code reasoning is an important capability of large language models. The reasoning ability of language models can be enhanced using chain-of-thought \citep{CoT}, RLVR \citep{O1, R1, lambert2025tulu}, and self-consistency \citep{wang2023selfconsistency}, in math \citep{MATH} and code \citep{LCB, USACO} domains. \citet{s1} and \citet{LIMO} focus on manually curating small-scale reasoning problems, which is sufficient to boost models' reasoning ability. More recently, \citet{openr1}, \citet{OCR}, and \citet{guha2025openthoughtsdatarecipesreasoning} have developed large-scale distillation methods from reasoning models to obtain high-quality long CoT SFT datasets that can be used to train student models effectively. \citet{nvidia2024nemotron4340btechnicalreport} curate reasoning datasets throughout the entire training pipeline. \citet{li2025codeio} introduce an innovative paradigm that transforms traditional code reasoning tasks from their original format into a ``given code + test cases / input-output prediction'' structure. Complementing these supervised learning approaches, \citet{deepcoder2025} demonstrate the effectiveness of reinforcement learning techniques applied to verified code reasoning problems. However, how to generate difficult synthetic coding problems and use them for training remains relatively underexplored.
\section{Conclusion}
In this paper, we propose a method for generating difficult code problems at scale. Specifically, we investigated a pipeline that uses majority voting to compute a proxy of difficulty and employs this as a signal for rejection fine-tuning of the problem generator, and combined it with novel difficulty-aware graph sampling prompts. This enables the trained generator to produce challenging problems at scale. We then use these generated problems for supervised fine-tuning (SFT) and reinforcement learning (RL) to verify their effectiveness. As a novel synthetic data generation method, we compared our approach with previous baselines at similar scales on code reasoning benchmarks and show that our method achieves better performance even when using less SFT data, particularly for hard problems.\\

\subsection*{Limitations and Future Work}
Although our method shows promise for rejection fine-tuning a generator, we still face limitations as the generator hasn't been trained using RL. One primary reason is that our current difficulty calculation is computationally expensive and challenging to implement in real-time to provide difficulty rewards in an RL pipeline, considering that we need to generate 8 responses and 20 test cases for each problem on the fly, execute them, and generate $K$ problems for each prompt. In future work, it would be worthwhile to explore methods that can provide rewards in real time, such as directly training a reward model to predict difficulty, or investigating other efficient approaches.

\newpage
\section*{Acknowledgments}
This work was done during the internship of HX at Microsoft Research. HX, JV, and RS acknowledge funding by the Swiss National Science Foundation (project InvestigaDiff; no. 10000503).
\section*{Reproducibility statement}
To help community reproduce our work, we described details of implementation in Section \ref{sec:implement}, which reports the details of data, benchmark, models, and hyperparameters we use in our experiments. We also report the framework we use for training and inference. In Appendix Figure  \ref{fig:promptgen} \ref{fig:prompttestcase} \ref{fig:promptconcept}, we report the prompt template we use.

\section*{Ethics statement}
In the paper, all the data we use is open-sourced. TACO \citep{taco} has Apache-2.0 license. LiveCodeBench \citep{LCB} and USACO \citep{USACO} are collected from open part of common competition websites.

\newpage

\bibliography{iclr2026_conference}
\bibliographystyle{iclr2026_conference}

\appendix
\newpage
\section{Appendix}

\begin{figure*}[h]
\vspace{4ex}
\makebox[\textwidth][l]{%
\includegraphics[width=.95\textwidth]{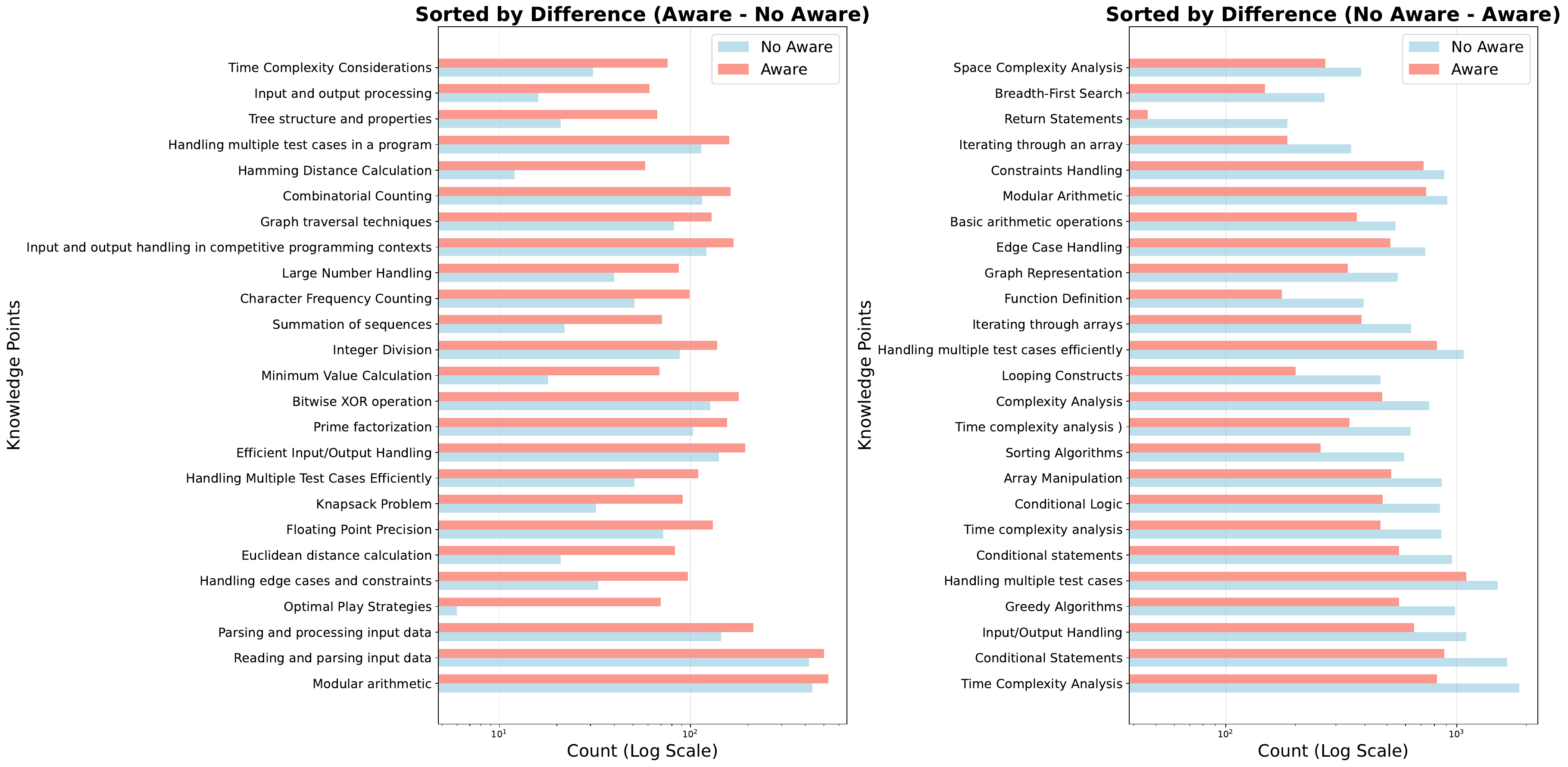}}
\caption{Examples of knowledge points where the count is increased (left) or decreased (right) when using difficulty-aware sampling.}
\vspace{1ex}
\label{fig:freq}
\end{figure*}

\subsection{Usage of LLM}
We used LLMs to polish the writing and assist in the implementation of parts of the codebase. All
the methodologies, ideas, conceptual contributions, experiment designs, analyses were developed solely by the authors.

\begin{table*}[]
\centering
\small
\caption{Case Study: Topic and knowledge points that were provided to the generator, and two generated questions. Left is generated by our trained model, right is generated by original Qwen2.5-14B-Instruct.}
\label{tab:case-study}
\begin{tabularx}{\textwidth}{>{\raggedright\arraybackslash}p{1.5cm} X X}
\toprule
Topic & 
\multicolumn{2}{>{\arraybackslash}p{\dimexpr\textwidth-3cm-4\tabcolsep-2\arrayrulewidth\relax}}{
String Analysis
} \\
\midrule
Knowledge points & 
\multicolumn{2}{>{\arraybackslash}p{\dimexpr\textwidth-3cm-4\tabcolsep-2\arrayrulewidth\relax}}{
1.~Hamming Distance Calculation; 2.~Time Complexity Optimization; 3.~Conditional Logic and State Management; 4.~Event Handling and Simulation; 5.~Priority Queue
} \\
\midrule
Problem & 
You are given a string s of length n consisting of binary characters (`0' and `1'). You need to perform several queries on this string. Each query can be one of the following types:\newline
1. set(l, r, c): Set all characters in the substring s[l:r+1] to the character c (either `0' or `1').\newline
2. flip(l, r): Flip all characters in the substring s[l:r+1] (change `0' to `1' and `1' to `0').\newline
3. hamming(l1, r1, l2, r2): Calculate the Hamming distance between the substrings s[l1:r1+1] and s[l2:r2+1].\newline
\newline
Your task is to handle these queries efficiently.\newline
\newline
Input:\newline
- The first line contains the string s of length n (1 $\leq$ n $\leq$ 200,000).\newline
- The second line contains an integer q (1 $\leq$ q $\leq$ 400,000) representing the number of queries.\newline
- The next q lines describe the queries. Each query is in one of the above formats. \newline
- For each hamming query, output the Hamming distance between the specified substrings.\newline
&
In a messaging system, messages are sent as strings of binary digits. Sometimes, due to technical issues, certain bits in the messages might get flipped (0 becomes 1, and 1 becomes 0). To ensure the integrity of the messages, a special protocol is used where each message is sent alongside its ``flipped'' version. Upon receiving both versions, the receiver calculates the Hamming distance between the original and the flipped message to detect any errors.\newline
\newline
Given two strings s and t of equal length representing the original and flipped versions of a message, your task is to write a function that calculates the Hamming distance between them. The Hamming distance is defined as the number of positions at which the corresponding bits are different. Additionally, the function should handle up to $10^5$ queries efficiently, optimizing for time complexity.\\
\bottomrule
\end{tabularx}
\end{table*}

\begin{figure*}[t]
\centering
\vspace{4ex}
\centerline{\includegraphics[width=.9\textwidth]{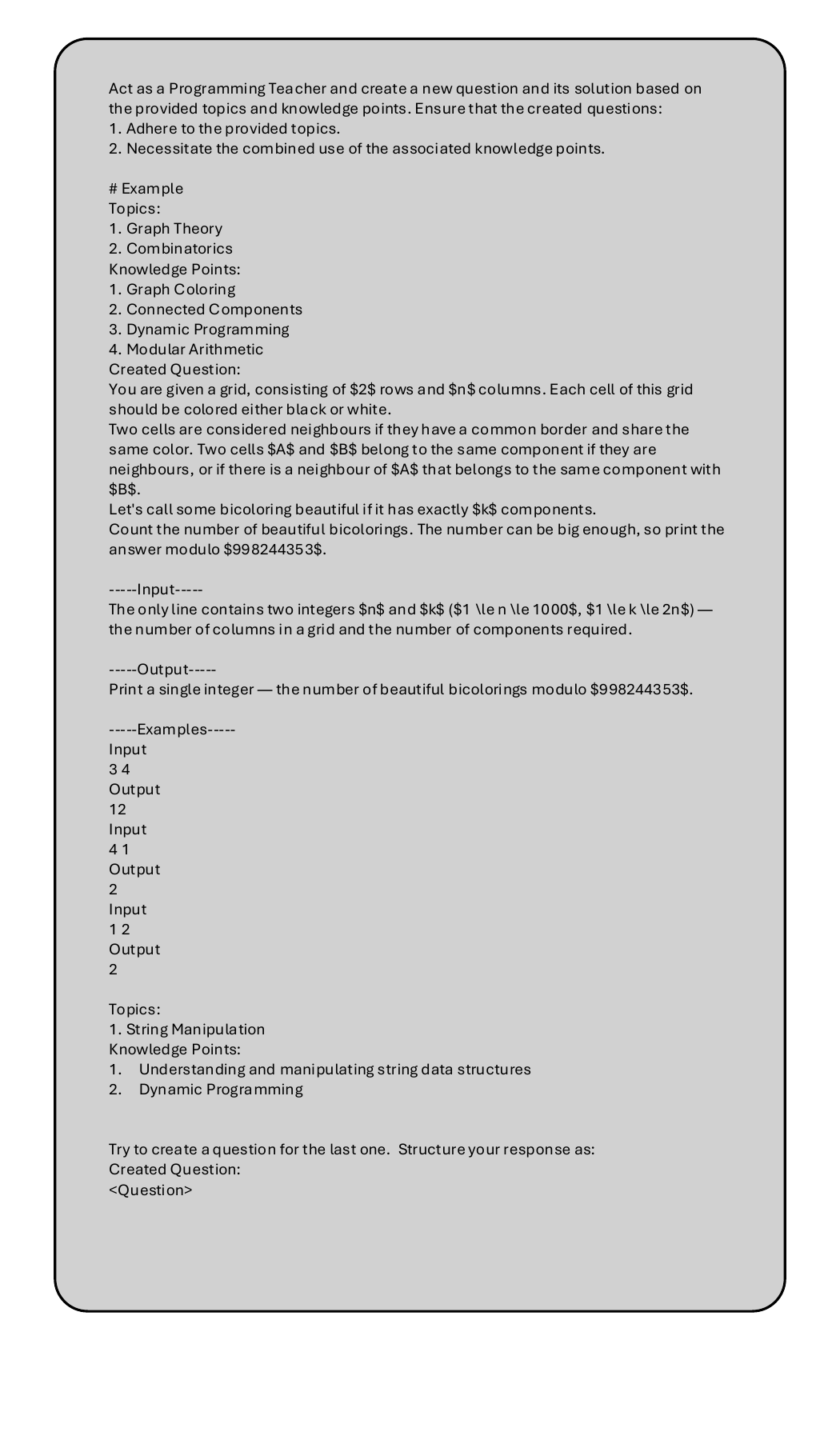}}
\caption{1-shot prompt example for problem generation. It is simplified for visualization, in real prompt, we have 8-shot for in-context learning. }
\vspace{1ex}
\label{fig:promptgen}
\end{figure*}

\begin{figure*}[t]
\centering
\vspace{1ex}
\centerline{\includegraphics[width=.9\textwidth]{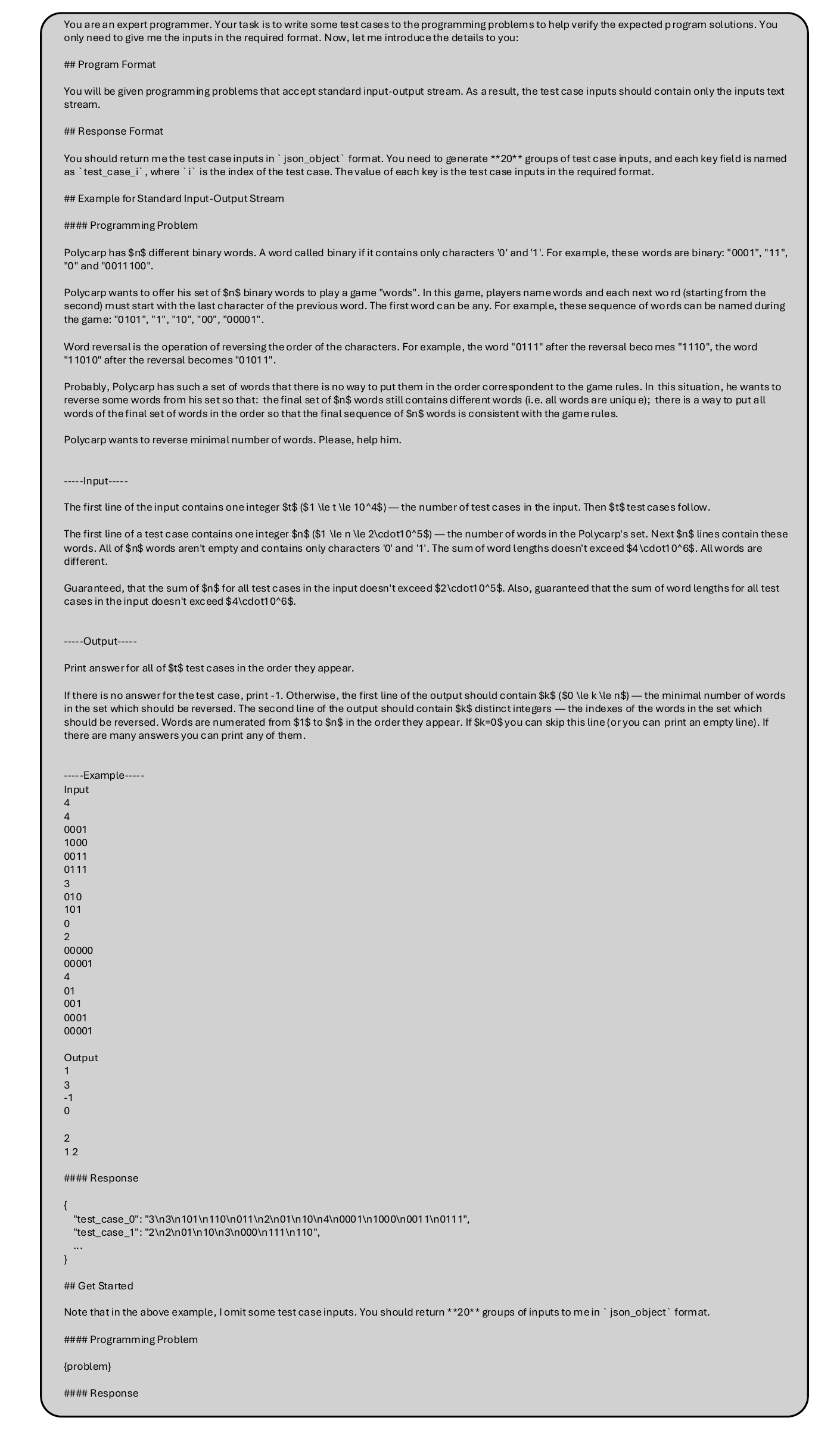}}
\caption{1-shot example prompt for testcase generation. }
\vspace{1ex}
\label{fig:prompttestcase}
\end{figure*}

\begin{figure*}[t]
\centering
\vspace{4ex}
\centerline{\includegraphics[width=.9\textwidth]{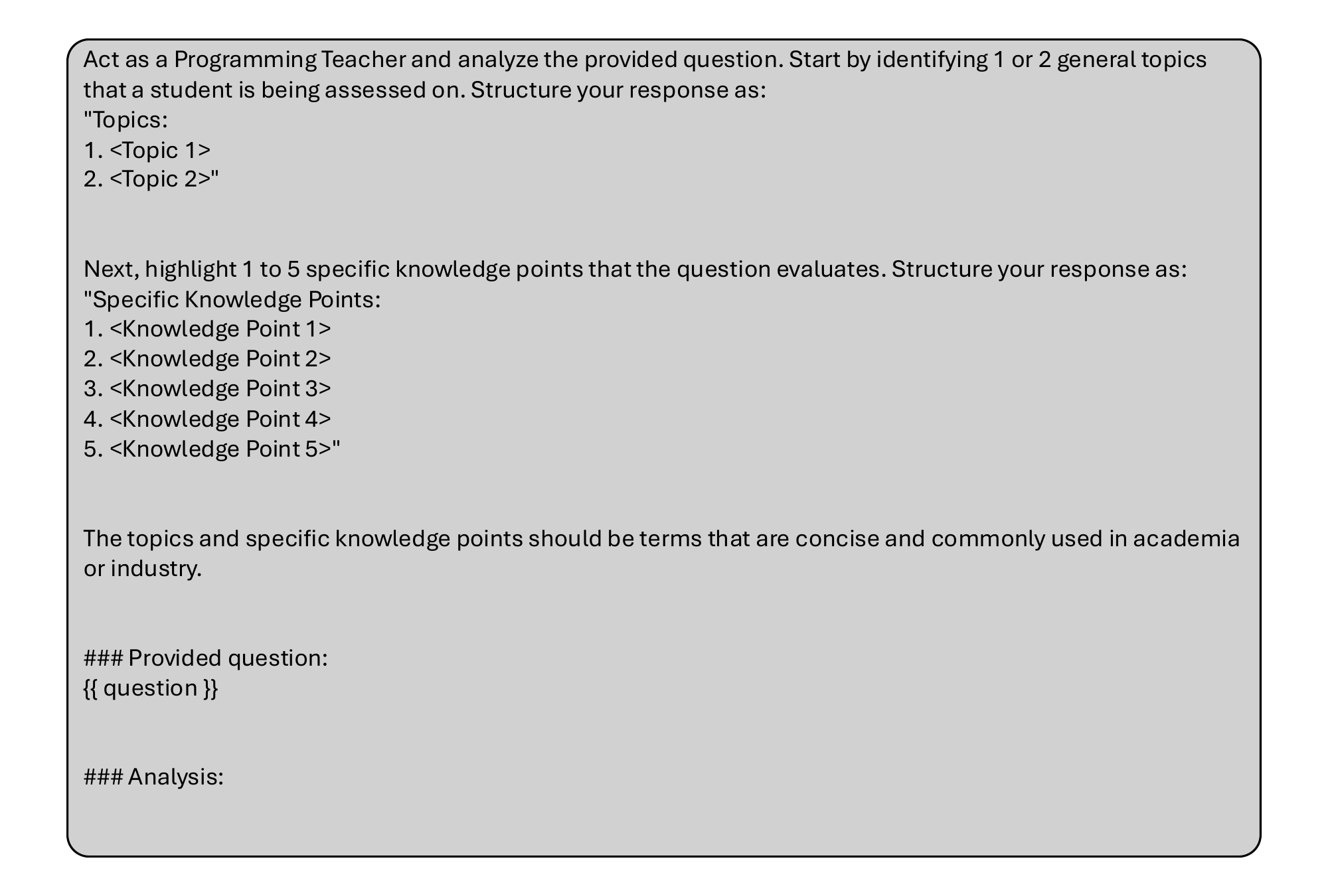}}
\caption{Prompt demonstration for concept extraction. }
\vspace{1ex}
\label{fig:promptconcept}
\end{figure*}

\end{document}